\documentclass[letterpaper]{article} 
\usepackage{aaai23}  
\usepackage{times}  
\usepackage{helvet}  
\usepackage{courier}  
\usepackage[hyphens]{url}  
\usepackage{graphicx} 
\usepackage{natbib}  
\usepackage{caption} 
\usepackage[frozencache,cachedir=mintedcache]{minted}
\usepackage{soul}
\usepackage{amsmath}
\usepackage{amssymb}
\usepackage{booktabs}
\usepackage{multirow}
\usepackage[table,xcdraw]{xcolor}
\usepackage[linesnumbered,ruled,vlined]{algorithm2e}
\usepackage{minted}
\DeclareMathOperator*{\argmin}{argmin}

\usepackage{newfloat}
\usepackage{listings}
\DeclareCaptionStyle{ruled}{labelfont=normalfont,labelsep=colon,strut=off} 
\floatstyle{ruled}
\newfloat{listing}{tb}{lst}{}
\floatname{listing}{Listing}
\title{CSTAR: Towards \underline{C}ompact and \underline{ST}ructured Deep Neural Networks with \underline{A}dversarial \underline{R}obustness}
\author{
    Huy Phan\textsuperscript{\rm 1}, Miao Yin\textsuperscript{\rm 1}, Yang Sui\textsuperscript{\rm 1}, Bo Yuan\textsuperscript{\rm 1}, Saman Zonouz\textsuperscript{\rm 2}
}
\affiliations{
    \textsuperscript{\rm 1}Department of Electrical and Computer Engineering, Rutgers University\\
    \textsuperscript{\rm 2}Schools of Cybersecurity and Privacy, Georgia Institute of Technology\\

}

\usepackage{bibentry}

\begin{document}

\maketitle

\begin{abstract}
  Model compression and model defense for deep neural networks (DNNs) have been extensively and individually studied. Considering the co-importance of model compactness and robustness in practical applications, several prior works have explored to improve the adversarial robustness of the sparse neural networks. However, the structured sparse models obtained by the existing works suffer severe performance degradation for both benign and robust accuracy, thereby causing a challenging dilemma between robustness and structuredness of compact DNNs.
  
  To address this problem, in this paper, we propose CSTAR, an efficient solution that simultaneously impose \textbf{C}ompactness, high \textbf{ST}ructuredness and high \textbf{A}dversarial \textbf{R}obustness on the target DNN models. By formulating the structuredness and robustness requirement within the same framework, the compressed DNNs can simultaneously achieve high compression performance and strong adversarial robustness. Evaluations for various DNN models on different datasets demonstrate the effectiveness of CSTAR. Compared with the state-of-the-art robust structured pruning, CSTAR shows consistently better performance. For instance, when compressing ResNet-18 on CIFAR-10, CSTAR achieves up to $20.07\%$ and $11.91\%$ improvement for benign accuracy and robust accuracy, respectively. For compressing ResNet-18 with $16 \times$ compression ratio on Imagenet, CSTAR obtains $8.58\%$ benign accuracy gain and $4.27\%$ robust accuracy gain compared to the existing robust structured pruning.
  
\end{abstract}

\section{Introduction}
\label{sec:intro}

Deep neural networks (DNNs) have obtained widespread popularity in various important intelligent systems, such as autonomous driving, smart home, computer-aided diagnosis, and so on. Despite their current unprecedented prosperity, DNNs are still facing two severe problems, namely \textit{efficiency challenge} and \textit{security challenge}, which can potentially hinder their future success in many practical applications. Specifically, the efficiency challenge refers to the high energy/storage/timing costs when deploying the modern large-scale DNN models in the field, especially on resource-constrained embedded and mobile devices. Meanwhile, the security challenge mainly refers to the vulnerability of DNN models to the attacks launched by the malicious adversary \cite{zang2020graph,xie2020real,phan2020cag,phan2022invisible,shi2022audio,phan2022ribac}, especially to the imperceptible and high-damage \textit{adversarial attacks}.  

To address these emerging challenges and further promote and democratize the use of DNNs, the machine learning community has conducted tremendous research activities to improve the execution efficiency and security of neural networks. In particular, \textit{model compression} and \textit{adversarial defense}, as two key techniques that can efficiently reduce the model size and enhance the model robustness, have been extensively studied in the existing literature. More specifically, various types of DNN compression approaches, e.g., pruning \cite{han2015learning,sui2021chip} and quantization \cite{gong2014compressing}, as well as many different DNN defense methods, e.g., adversarial training \cite{madry2017towards,wong2020fast} and input denoising \cite{prakash2018deflecting}, have been developed in both academia and industry.

\textbf{Co-exploring Model Compactness \& Robustness.} Considering the co-importance of compactness and robustness for the DNNs used in the practical applications, an attractive and natural thought is to develop a method that can generate neural networks with high compactness \textit{and} strong robustness simultaneously. Motivated by this idea, several prior works have explored to examine and further improve the robustness of the compressed DNNs. In particular, because of the popularity of \textit{pruning} in many model compression tasks, most of the existing compactness/robustness co-exploration efforts \cite{sehwag2020hydra,sehwag2019towards,ye2019adversarial,vemparala2021adversarial} focus on  efficient approaches to generate the robust pruned DNN models. To be specific, \cite{ye2019adversarial} demonstrates that the model robustness and compactness can co-exist for neural networks via concurrent adversarial training and weight pruning. Considering the pruning criterion in \cite{sehwag2019towards} is designed to preserve benign accuracy instead of improving robust accuracy, \cite{sehwag2020hydra} further proposes robust training-aware pruning to enhance both.

\textbf{The Dilemma Between Robustness and Structuredness of Compact DNNs.} Although the state-of-the-art robust pruning works can simultaneously enable high sparsity and strong robustness for the target DNN models, all of their best performance comes from performing \textit{unstructured pruning}. As extensively reported and observed in the prior literature \cite{han2016eie,liu2015sparse,zhu2020efficient,deng2021gospa}, the unstructured sparse models, though can indeed exhibit high accuracy and high compression ratio, cannot bring considerable speedup on the off-the-shelf CPU/GPU. To mitigate this challenge, based on the observation that the structured pruning, e.g., channel/filter pruning, can enable the measurable acceleration, prior efforts have also explored the robustness-aware structured pruning to strike for simultaneous robustness, compactness and structuredness of DNN models. Unfortunately, as reported in \cite{sehwag2019towards,sehwag2020hydra,ye2019adversarial},  \ul{the existing robust structured sparse models suffer severe performance degradation as compared to their unstructured counterparts with respect to both benign and robust accuracy.} Consequently, the challenging dilemma between robustness and structuredness of the compact DNNs has not been well addressed yet.

\textbf{Low-rank Robustness: A Structured Method Towards Compact \& Robust DNNs.} From the perspective of model compression, structured pruning is not the only way to produce structured compact models. \textit{Low-rank tensor decomposition} \cite{denton2014exploiting,kim2015compression,yang2017tensor,pan2019compressing,yin2021towards,yin2022hodec}, as another type of popular model compression technique, can also ensure that the compressed model exhibits the desired high structuredness for the practical speedup. To be specific, different from structured pruning that removes the entire channels/filters to enable the structured sparsity, low-rank tensor decomposition explores the structure-level model redundancy and factorizes the original dense model into multiple tensor cores. Because those decomposed tensor cores are still in the dense format and their involved computations are entirely based on the general matrix multiplication (GEMM), considerable acceleration can be observed and measured on the off-the-shelf CPU/GPU \cite{kim2015compression,zhang2015efficient,zhang2015accelerating}. Evidently, such hardware-friendly low-rankness-based compactness, if can further co-exist with the adversarial robustness, will be attractive for compact and robust DNN model design.

\textbf{Questions to be Answered.} Despite the promising potentials of robust tensor decomposed models, several important questions need to be answered and addressed. To be specific, what is the suitable perspective to connect and unify the low-rankness and robustness of a DNN model? Once we can formulate a unified problem, what is the efficient approach to simultaneously impose low-rankness and robustness with a high compression ratio, benign accuracy, and robust accuracy? Also, considering the conventional tensor decomposition requires a tedious manual rank selection process, which will be very time-consuming if integrated with defense methods such as adversarial training, is there any efficient automatic tensor rank determination scheme?

\textbf{Technical Preview and Contributions.} To answer these questions and deliver the promise of robust low-rank DNN models, in this paper we study and develop an efficient solution, namely CSTAR, that can simultaneously impose the desired high low-rankness-based \textbf{C}ompactness, high \textbf{ST}ructuredness and high \textbf{A}dversarial \textbf{R}obustness on the target DNN models. By formulating the low-rankness and robustness requirement within the same framework as well as determining the tensor ranks in a global way, the compressed DNNs can achieve high compression performance and strong adversarial robustness at the same time. Overall, the contributions of this paper are summarized as follows:

\begin{itemize}
    \item We propose a framework that formulates the low-rankness and adversarial robustness requirement to a unified constrained optimization problem. Based on that, we then propose an adversarial training process that can gradually impose the desired low-rankness on the DNN model, thereby simultaneously ensuring high compression performance and adversarial robustness.
    
    \item We propose a low-cost automatic rank selection scheme that can select the desired tensor ranks for each layer in a very convenient way. Such selection can be dynamically adjusted during the compression procedure, thereby avoiding the tedious manual rank selection and meanwhile ensuring high performance.
    
    \item We evaluate the performance of the CSTAR solution for various DNN models on different datasets. Compared with the state-of-the-art robust structured pruning methods, our proposed approach shows consistently better performance. For instance, when compressing ResNet-18 on the CIFAR-10 dataset, CSTAR can achieve up to $20.07\%$ and $11.91\%$ increase for benign accuracy and robust accuracy, respectively. On the Imagenet dataset, for compressing ResNet-18 with $16 \times$ compression ratio, CSTAR can obtain $8.58\%$ benign accuracy gain and $4.27\%$ robust accuracy gain as compared to the existing robust structured pruning method.
\end{itemize}

\section{Background and Preliminaries}
\label{sec:preliminary}

\subsection{Tensor Decomposition for Model Compression}
\label{subsec:tensor_decomp}

Tensor decomposition, as a type of low-rank decomposition approach, has been widely used in many DNN model compression works. According to tensor theory, there exist various types of tensor decomposition approaches, such as Tucker \cite{tucker1963implications}, CP \cite{harshman1970foundations}, Tensor Train \cite{oseledets2011tensor} and Tensor Ring \cite{zhao2016tensor}. Without loss of generality, in this paper, we adopt Tucker-2 decomposition \cite{kim2015compression} as the underlying low-rank tensor method for model compression. 

\textbf{Tensor Contraction.} Tucker-2 decomposition can be compactly represented via using tensor contraction. In general, tensor contraction can be performed between any two tensors that have at lease one matched dimension. For instance, consider two tensors $\boldsymbol{\mathcal{A}} \in \mathbb{R}^{d_1 \times d_2 \times n}$ and $\boldsymbol{\mathcal{B}} \in \mathbb{R}^{n \times d_3 \times d_4}$, the output tensor $\boldsymbol{\mathcal{C}}$ of size $\mathbb{R}^{d_1 \times d_2 \times d_3 \times d_4}$ after the tensor contraction can be calculated as follow:

\begin{equation}
\boldsymbol{\mathcal{C}}_{(a_1, a_2, b_1, b_2)} = \boldsymbol{\mathcal{A}} \times^3_1 \boldsymbol{\mathcal{B}} = \sum_{i=1}^n  \boldsymbol{\mathcal{A}}_{(a_1, a_2, i)} \boldsymbol{\mathcal{B}}_{(i, b_1, b_2)}.
\end{equation}

\textbf{Tucker-2 Tensor Decomposition.} With the notation of tensor contraction, given the weight tensor of convolutional layer $\boldsymbol{\mathcal{W}}\in\mathbb{R}^{O\times I\times K\times K}$ as well as the corresponding input tensor $\boldsymbol{\mathcal{X}}\in\mathbb{R}^{I\times H\times W}$ and output tensor $\boldsymbol{\mathcal{Y}}\in\mathbb{R}^{O \times H'\times W'}$, $\boldsymbol{\mathcal{W}}$ can be factorized via Tucker-2 decomposition as:

\begin{equation}
\boldsymbol{\mathcal{W}} = \boldsymbol{\mathcal{U}}^1 \times^2_2 ~ \boldsymbol{\mathcal{U}}^2  \times^2_2 ~ \boldsymbol{\mathcal{G}},
\label{eqn:decompose_w}
\end{equation}
where $\boldsymbol{\mathcal{U}}^1 \in\mathbb{R}^{O \times R_1}$, $\boldsymbol{\mathcal{U}}^2 \in \mathbb{R}^{I \times R_2}$, and $\boldsymbol{\mathcal{G}} \in \mathbb{R}^{R_1 \times R_2 \times K\times K}$. $R_1$ and $R_2$ denote the Tucker-2 tensor ranks. Then, the output tensor $\boldsymbol{\mathcal{Y}}=\boldsymbol{\mathcal{U}}^1 \times^2_1 \boldsymbol{\mathcal{T}}^2$ can be calculated via tensor contraction-based computations:

\begin{align}
    \boldsymbol{\mathcal{T}}^1 = \boldsymbol{\mathcal{U}}^2 \times^1_1 \boldsymbol{\mathcal{X}}, ~~~ \boldsymbol{\mathcal{T}}^2 = \sum_{p=1}^{K}\sum_{q=1}^{K}\boldsymbol{\mathcal{G}}_{(:,:,p,q)} \times^2_1 \boldsymbol{\mathcal{T}}^1
\label{eqn:convolution}
\end{align}
where $\boldsymbol{\mathcal{T}}^{1} \in \mathbb{R}^{R_2 \times H \times W}$, and $\boldsymbol{\mathcal{T}}^{2} \in \mathbb{R}^{R_1 \times H' \times W'}$ are the intermediate results. Notice that tensor contraction has been well supported by PyTorch and TensorFlow platforms via using \texttt{torch.tensordot} and \texttt{tf.tensordot}.

\subsection{Adversarial Training for DNN Defense}
\label{sec:pgd_and_at}

\textbf{Projected Gradient Decent (PGD).} PGD is a very popular adversarial examples generation approach \cite{madry2017towards} because of its high-quality and fast generation process. In general, PGD first calculates the gradient w.r.t. the benign input data $x$ and then updates the input data in the direction that maximizes the loss function $\mathcal{L}$. The resulting adversarial example $x_{adv}$ is then projected to the maximum allowed perturbation $\Delta$ via using projection operator $\boldsymbol{\Pi}$. Such process is repeated $k$ times to obtain a strong example:

\begin{align}
    \boldsymbol{\mathcal{X}}_{adv}^{N+1} & = \boldsymbol{\Pi}_\Delta(\boldsymbol{\mathcal{X}}_{adv}^{N} + \epsilon \cdot \text{sign}(\nabla  \mathcal{L}(\boldsymbol{\mathcal{W}}, \boldsymbol{\mathcal{X}}_{adv}^{N}, \boldsymbol{y})),
\label{eqn:pgd}
\end{align}
where $\boldsymbol{\mathcal{X}}_{adv}^{N+1}$ is the set of adversarial examples at the $N$-th iteration, $\epsilon$ is the step size, and $\boldsymbol{y}$ is set of the labels.

\textbf{Adversarial Training.}
Adversarial training (AT) \cite{madry2017towards} is one of the most effective methods to protect DNNs against adversarial attacks. The key idea of adversarial training is to generate adversarial examples during the training process to make the model can better fit for adversarial examples. In general, adversarial training aims to solve the following min-max optimization problem:

\begin{equation}
    \min_{\boldsymbol{\mathcal{W}}} ~ \max_{\delta\in \boldsymbol{\Delta}} ~ \mathcal{L}(\boldsymbol{\mathcal{W}}, \boldsymbol{\mathcal{X}}+\delta, \boldsymbol{y}),
\end{equation}
where $\boldsymbol{\mathcal{X}}$ denotes the set of benign inputs, and $\mathcal{L}$ is the loss function. Here the inner maximization problem can be solved by the above described PGD.

\section{Related Works}
\label{sec:related_works}

\textbf{Model Compression with Adversarial Training.} To date several prior works have explored the efficient integration of model compression to the adversarial training process. To be specific, \cite{sehwag2019towards,ye2019adversarial,sehwag2020hydra,vemparala2021adversarial,xie2020blind,guo2018sparse,madaan2020adversarial,rakin2019robust,ozdenizci2021training} investigate the robustness-aware pruning to achieve high model sparsity and adversarial robustness. Also, \cite{fu2021double,lin2019defensive} propose several approaches to develop low-bit precision robust models. In addition, \cite{gui2019model} studies the adversarial robust model with the combination of sparsity, low bit-precision and matrix factorization. However, most of these works are built either on unstructured sparsity or complicated hybrid-precision schemes (e.g, 4-8 bits), which can only be supported by the specialized hardware, thereby severely limiting their practical deployment on the off-the-shelf CPUs/GPUs. To date, only \cite{sehwag2019towards,ye2019adversarial,sehwag2020hydra} provide the structured pruning solutions for adversarial robust DNN models that can exhibit practical speedup. However, the corresponding  benign and robust accuracy is not satisfied.

\textbf{Low-rank Matrix \& Tensor Decomposition.} Similar to structured pruning, low-rank decomposition is another powerful structuredness-ensuring compression approach that can bring considerable speedup. In general, low-rank decomposition can be realized via either matrix decomposition or tensor decomposition. \textbf{Matrix decomposition} \cite{tai2015convolutional,li2018constrained,xu2019trained,yang2020learning,idelbayev2020low,xu2019trained} uses SVD to factorize the large matrix to small matrix components. In particular, when the objective is the 4-D weight tensor of convolutional layer, it has to first reshapes the 4-D tensor to 2-D matrix, and then performs decomposition to obtain small 2-D matrices. \ul{Evidently, such flattening strategy can not fully exploit the inherent spatial information of the 4-D weight tensor, thereby causing limited performance.} Motivated by these limitations, \textbf{Tensor decomposition}, such as Tucker \cite{kim2015compression}, Tensor Train \cite{oseledets2011tensor} and Tensor Ring \cite{zhao2016tensor}, \ul{which can directly factorize the original high-order tensor in the high-dimensional space}, have been studied to facilitate DNN compression. Because tensor decomposition can fully exploit and leverage the rich spatial correlation and information of weight tensor in the high-order tensor space, it can bring impressive compression performance \cite{tai2015convolutional,kim2015compression,li2018constrained,wang2018wide}

\textbf{Rank Selection.} Determining the proper rank is very critical for low-rank decomposed DNN models. To date, most of the existing low-rank DNN compression approaches \cite{kim2015compression,zhao2016tensor,yang2017tensor,xu2019trained,idelbayev2020low,yin2021towards2} select the rank in a manual way. However, because each layer needs to be assigned with one or more rank values, such heuristic searching is a very tedious and time-consuming procedure. Recently, some works have begun to study efficient automatic rank determination. In \cite{li2021heuristic,hajimolahoseinicompressing}, progressive search-based methods are proposed to identify proper low matrix rank \cite{hajimolahoseinicompressing} and low tensor rank \cite{li2021heuristic}. However, this strategy is performance-limited because it still partially needs manual selection \cite{hajimolahoseinicompressing} or costly iterative searching and sampling \cite{li2021heuristic}. In addition, \cite{gusak2019automated,ye2018learning} propose to automatically select the rank in a layer-wise way. However, such strategy is not the ideal solution because of the inevitable high cost: \cite{gusak2019automated} needs expensive Bayesian calculation plus setting additional auxiliary hyper-parameters during the searching, and \cite{ye2018learning} needs many rounds of enumeration to search for the proper rank values. Moreover, determining the rank layer by layer, in principle, is not a promising solution since it may not bring the globally optimal rank determination for the entire model.

\textbf{Advantages of Our Proposed CSTAR.} Compared with the above discussed related works, CSTAR enjoys the following benefits:
\begin{itemize}
    \item \textbf{Practical Measurable Speedup.} Compared with most existing works on joint compression and adversarial training via unstructured pruning \cite{ye2019adversarial,sehwag2020hydra}, low-bit precision \cite{fu2021double,lin2019defensive} or hybrid combination \cite{gui2019model}, CSTAR can bring practical speedup on the off-the-shelf CPUs/GPUs because of its strong model structuredness provided by low-rankness (see Section \ref{sec:experiments}).
    \item \textbf{Low-Cost Automatic Rank Selection.} The rank selection of CSTAR is naturally integrated to the compression process and automatically learned from data, and hence it has very low cost without any expensive enumeration or searching operation that prior works \cite{li2021heuristic,hajimolahoseinicompressing,gusak2019automated} need. Meanwhile, the ranks for all the layers are jointly determined in a global way (\ul{see Appendix}).
    \item \textbf{High Benign and Robust Accuracy.} Compared with the state-of-the-art structured pruning-based adversarial training methods \cite{ye2019adversarial,sehwag2020hydra} that can also bring practical speedup, CSTAR shows very significant performance improvement in terms of benign accuracy and robust accuracy across different models on various datasets (see Section \ref{sec:experiments}).
\end{itemize}

\section{Our Proposed Method}
\label{sec:method}

\textbf{Problem Formulation.} CSTAR aims to co-explore adversarial robustness and low-rankness on DNN models to achieve high compression ratio, high benign accuracy, and high robust accuracy with practical speedup simultaneously. To achieve this goal, a suitable perspective that can properly connect and unify the adversarial robustness and low-rankness should be identified. Considering adversarial training is currently the most important and popular adversarial defense approach, we formulate our goal as the following optimization problem:

\begin{align}
\begin{split}
&\min_{\boldsymbol{\mathcal{W}}_i} ~ \max_{\delta\in \boldsymbol{\Delta}} ~ \mathcal{L}(\boldsymbol{\mathcal{W}}, \boldsymbol{\mathcal{X}}+\delta, \boldsymbol{y}), \\
&\text{s.t.} ~~~ \texttt{rank}(\boldsymbol{\mathcal{W}}_i) \le \textbf{t}_i, ~~~ i = 1, \dots, N,
\end{split}
\label{eqn:problem_formulation}
\end{align}
where $N$ denotes the number of layers, $\texttt{rank}(\cdot)$ is the function that calculates the Tucker-2 ranks $\textbf{r} = [r_1, r_2]$ of the input tensor, and $\textbf{t} = [t_1, t_2]$ denotes the target ranks that can satisfy the compression ratio. Here $\textbf{r} \le \textbf{t}$ operation means element-wise comparison.

\textbf{Challenges to be Addressed.} In general, solving problem \ref{eqn:problem_formulation} is non-trivial yet challenging. More specifically, two critical issues need to be resolved:

\begin{figure*}[ht]
\centering
\includegraphics[width=0.825\textwidth]{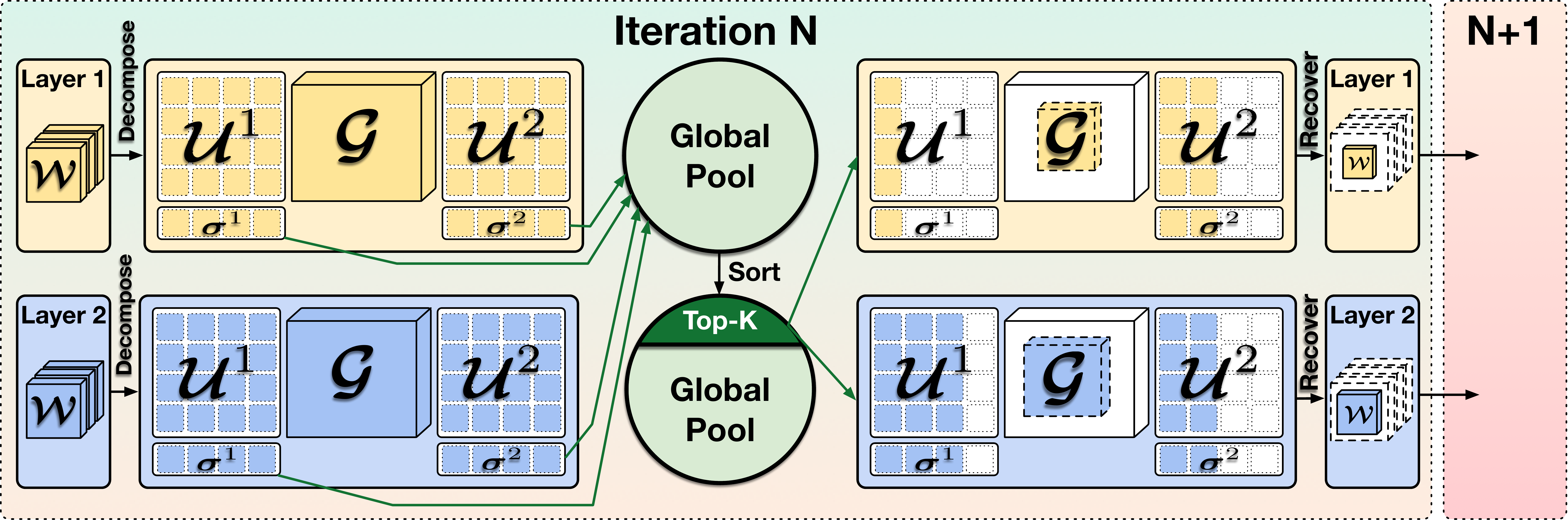}
\caption{Proposed singular value-based automatic tensor rank selection. Here the rank selection can vary as training progresses.}
\label{fig:global_rank}
\end{figure*}

\textbf{\ul{Challenge \#1:}} \textit{How should we optimize problem \ref{eqn:problem_formulation} that is constrained by the non-differentiable $\texttt{rank}(\boldsymbol{\mathcal{W}}_i) \le \textbf{t}_i$?}

\ul{Analysis.} A straightforward way is to directly decompose the pre-trained $\{\boldsymbol{\mathcal{W}}_i\}$ into the suitable low-rank format that satisfies the rank constraint, and then fine-tune the entire decomposed model. Though this direct strategy can indeed bring the desired low-rankness, such straightforward and explicit decomposition on the high-accuracy model will probably cause considerable approximation error. Consider that the common pre-trained CNN models typically lack sufficient low-rank property, the corresponding approximation error and its resulting benign and robust accuracy loss, consequently, will be too huge to be recovered after fine-tuning. 

\begin{algorithm}[t]
\SetAlgorithmName{Alg.}{}{}
\begin{minted}[
linenos,
% frame=lines,
% framesep=2mm,
baselinestretch=0.95,
fontsize=\footnotesize]{python}
def decompose(W, r1, r2):
#Input: Weight tensor W, ranks r1, r2.
#Ouput: Decomposed U1, U2, G tensors.
  W = W.flatten(start_dim=2)
  t1 = W.flatten(start_dim=1)
  U1, sigma1, V1 = torch.svd(t1)
  t2 = W.permute(1,0,2).flatten(start_dim=1)
  U2, sigma2, V2 = torch.svd(t2)
  G = torch.tensordot(U2,W,dims=([0],[1]))
  G = torch.tensordot(U1,G,dims=([0],[1]))
  U1,U2,G = U1[:,:r1], U2[:,:r2], G[:r1,:r2]
  return U1, U2, G
def recover(U1, U2, G):
#Input: Decomposed U1, U2, G tensors.
#Output: Recovered low-rank weight tensor W.
  W = torch.tensordot(U2,G,dims=([1],[1]))
  W = torch.tensordot(U1,W,dims=([1],[1]))
  O, I, KK = W.size()
  K = int(KK ** 0.5)
  return W.view(O, I, K, K)
\end{minted}

\caption{Tucker-2 projection for solving Eqn. \ref{eqn:z_update}}
\label{algorithm1}
\end{algorithm}

\ul{Our Proposed Solution.} To address this challenge, \ul{we propose to gradually impose the desired low-rankness on the DNN model during the adversarial training procedure.} To be specific, the to-be-compressed model is first trained in the full-rank format but gradually exhibits low-rank characteristic as well as adversarial robustness. Our rationale for this indirect training strategy lies in the belief that the smooth and ``soft" transition from full-rank to low-rank format, intuitively, can better preserve benign and robust accuracy than the direct and ``hard" decomposition. Motivated by this philosophy, we re-formulate problem \ref{eqn:problem_formulation} via adding the low-rankness requirement as a \textbf{soft regularization term}:

\begin{align}
\begin{split}
\min_{\boldsymbol{\mathcal{W}}_i, \boldsymbol{\mathcal{Z}}_i \in \mathcal{S}} & ~ \max_{\delta\in \boldsymbol{\Delta}} ~ \mathcal{L}(\boldsymbol{\mathcal{W}}_i, \boldsymbol{\mathcal{X}}+\delta, \boldsymbol{y}),\\
\text{s.t.} ~~~ \boldsymbol{\mathcal{W}}_i & = \boldsymbol{\mathcal{Z}}_i, ~~~ i = 1, \dots, N,
\end{split}
\end{align}
where $\boldsymbol{\mathcal{Z}}_i$ is the introduced auxiliary variables and $\mathcal{S} = \{\boldsymbol{\mathcal{W}}_i ~|~ \texttt{rank}(\boldsymbol{\mathcal{W}}_i) \le \textbf{t}_i \}$. Then, we further reformulate the problem via using the augmented Lagrangian:

\begin{equation}
\label{eq:augment}
\begin{split}
\min_{\boldsymbol{\mathcal{W}}_i, \boldsymbol{\mathcal{Z}}_i \in \mathcal{S}} ~ \max_{\boldsymbol{\mathcal{M}}_i, \delta\in \boldsymbol{\Delta}} ~ \mathcal{L}(\boldsymbol{\mathcal{W}}_i, \boldsymbol{\mathcal{X}}+\delta, \boldsymbol{y}) \\ +  \frac{\rho}{2} \sum_{i=1}^{N} \left ( \left \| \boldsymbol{\mathcal{W}}_i - \boldsymbol{\mathcal{Z}}_i + \boldsymbol{\mathcal{M}}_i \right \|_{F}^{2} + \| \boldsymbol{\mathcal{M}}_i \|_F^2 \right ),
\end{split}
\end{equation}
where $\boldsymbol{\mathcal{M}}_i$ are dual multipliers and $\rho > 0$ is the augmented parameter. After that, problem \ref{eq:augment} can be solved via Dykstra’s alternating projection method \cite{bauschke1994dykstra}, a primal-dual optimization algorithm that is well-suited for multi-term minimization. To be specific, the iterative updates are performed as:

\textbf{Update $\{\boldsymbol{\mathcal{W}}_i\}$ using SGD.}
In this step, we optimize the sub-problem as:

\begin{equation}
\min_{\{\boldsymbol{\mathcal{W}}_i\}} \underbrace{\max_{\delta\in \boldsymbol{\Delta}} \mathcal{L}(\{\boldsymbol{\mathcal{W}}_i\}, \boldsymbol{\mathcal{X}}+\delta, \boldsymbol{y})}_\text{Adversarial Training} + \underbrace{\frac{\rho}{2} \sum_{i=1}^{N} \left \| \boldsymbol{\mathcal{W}}_i - \boldsymbol{\mathcal{Z}}_i + \boldsymbol{\mathcal{M}}_i \right \|_{F}^{2}}_\text{Low-rank Regularization}.
\label{eqn:sub_w_reformulate}
\end{equation}

The first term in Eq. \ref{eqn:sub_w_reformulate} is to find adversarial examples, and it can be realized via using PGD described in Sec. \ref{sec:pgd_and_at}. The second term is to gradually impose low-rank property onto $\{\boldsymbol{\mathcal{W}}_i\}$. Since both terms are differentiable, we can iteratively solve Eq. \ref{eqn:sub_w_reformulate} using stochastic gradient descent (SGD) with learning rate $\alpha$:

\begin{align}
\begin{split}
\boldsymbol{\mathcal{W}}_i \gets \boldsymbol{\mathcal{W}}_i & - \alpha \cdot \nabla_{\boldsymbol{\mathcal{W}}_i} \left[ \mathcal{L}(\boldsymbol{\mathcal{W}}_i, \boldsymbol{\mathcal{X}}+\delta, \boldsymbol{y}) \right] \\ & - \alpha \cdot \rho \left(  \| \boldsymbol{\mathcal{W}}_i - \boldsymbol{\mathcal{Z}}_i + \boldsymbol{\mathcal{M}}_i ) \right).
\end{split}
\label{eqn:w_update}
\end{align}

\textbf{Update $\{\boldsymbol{\mathcal{Z}}_i\}$ using Projection.} In this step, $\{\boldsymbol{\mathcal{Z}}_i\}$ can be updated as:

\begin{equation}
\begin{aligned}
\boldsymbol{\mathcal{Z}}_i \gets \argmin_{\boldsymbol{\mathcal{Z}}_i \in \mathcal{S}} ~ \left \| \boldsymbol{\mathcal{W}}_i - \boldsymbol{\mathcal{Z}}_i + \boldsymbol{\mathcal{M}}_i \right \|_{F}^{2}.
\end{aligned}
\label{eqn:min_z}
\end{equation}

More specifically, problem \ref{eqn:min_z} can be solved via projecting $\boldsymbol{\mathcal{Z}}_i$ onto set $\mathcal{S}$ as:

\begin{equation}
\begin{aligned}
\boldsymbol{\mathcal{Z}}_i \gets \textbf{Prj}(\boldsymbol{\mathcal{W}}_i + \boldsymbol{\mathcal{M}}_i),
\end{aligned}
\label{eqn:z_update}
\end{equation}
where the projection truncates the ranks of $\boldsymbol{\mathcal{W}}_i + \boldsymbol{\mathcal{M}}_i$ to the desired target ranks $\textbf{t}_i$. Algorithm \ref{algorithm1} describes the details of this low-rank projection. 

\textbf{\ul{Challenge \#2:}} \textit{How should we properly select the tensor ranks in a convenient way?}

\begin{table*}[ht]
\setlength{\tabcolsep}{10pt}
\centering
\scalebox{0.8}{
\begin{tabular}{
>{\columncolor[HTML]{FFFFFF}}l 
>{\columncolor[HTML]{FFFFFF}}c
>{\columncolor[HTML]{FFFFFF}}c 
>{\columncolor[HTML]{FFFFFF}}c 
>{\columncolor[HTML]{FFFFFF}}c 
>{\columncolor[HTML]{EFEFEF}}c}
\toprule
Comp. & \textbf{LWM} & \textbf{LWM} & \textbf{RAP} & \textbf{RAP} & \textbf{CSTAR} \\
Ratio & Filter Pruning & Column Pruning & Filter Pruning & Column Pruning & Low-Rank Decomp. \\ \midrule
\multicolumn{6}{c}{\textbf{VGG-16 (Pre-train 80.08 / 44.10)}} \\
$2\times$ & 73.69 / 44.05 & 77.31 / 44.96 & 76.85 / 46.32 & 79.25 / 45.30 & \textbf{79.71 / 47.23} \\
$4\times$ & 63.76 / 38.93 & 70.46 / 41.54 & 71.48 / 42.59 & 75.55 / 43.97 & \textbf{79.64 / 47.06} \\
$8\times$ & 54.30 / 34.45 & 63.69 / 38.35 & 62.48 / 37.58 & 69.54 / 41.52 & \textbf{79.23 / 46.42} \\
$16\times$ & 45.77 / 30.37 & 55.29 / 35.01 & 53.72 / 33.72 & 60.55 / 37.36 & \textbf{77.82 / 46.13} \\
$32\times$ & 40.17 / 27.35 & 47.94 / 31.71 & 43.78 / 29.34 & 52.62 / 34.03 & \textbf{75.82 / 45.43} \\
$64\times$ & 10.00 / 10.00 & 40.01 / 29.35 & 10.00 / 10.00 & 45.08 / 32.49 & \textbf{70.77 / 43.13} \\
\addlinespace[0.5em]
\multicolumn{6}{c}{\textbf{ResNet-18 (Pre-train 83.23 / 49.78)}} \\
$2\times$ & 83.62 / 45.47 & 83.57 / 44.91 & 82.68 / 48.85 & 82.51 / 49.45 & \textbf{83.86 / 50.06} \\
$4\times$ & 79.10 / 46.49 & 80.54 / 46.79 & 79.78 / 46.78 & 81.81 / 46.76 & \textbf{83.22 / 49.48} \\
$8\times$ & 70.16 / 42.41 & 75.23 / 44.46 & 71.82 / 43.17 & 76.89 / 45.57 & \textbf{82.30 / 48.81} \\
$16\times$ & 61.31 / 37.12 & 68.55 / 40.86 & 64.39 / 38.01 & 70.18 / 41.78 & \textbf{80.90 / 47.36} \\
$32\times$ & 51.59 / 32.43 & 59.20 / 36.60 & 54.31 / 33.32 & 62.68 / 37.15 & \textbf{77.73 / 46.47} \\
$64\times$ & 42.36 / 28.52 & 50.92 / 32.16 & 46.97 / 29.41 & 52.01 / 32.48 & \textbf{72.08 / 44.39} \\
\bottomrule
\end{tabular}}
\caption{Benign accuracy / PGD-50 robust accuracy (\%) on CIFAR-10 with Standard AT.}
\label{table:sota}
\end{table*}

\begin{algorithm}[t]
\SetAlgorithmName{Alg}{}{}
\textbf{Input:} {Dataset $\boldsymbol{\mathcal{D}}$, pre-trained weight $\boldsymbol{\mathcal{W}}$, comp. ratio $r$, low-rank reg. steps $T_1$, fine-tune steps $T_2$}.\\
\textbf{Output:} {Fine-tuned $\{\boldsymbol{\mathcal{U}}^1_i, \boldsymbol{\mathcal{U}}^2_i, \boldsymbol{\mathcal{G}}_i\}$}. \\
$\{\boldsymbol{\mathcal{Z}}_i\} \gets \{\boldsymbol{\mathcal{W}}_i\}$; $\{\boldsymbol{\mathcal{M}}_i\} \gets \textbf{0}$;\\
\For ({~~~~~\textcolor{brown}{\textit{$\triangleright$ low-rank regularization}}}) {$t=1$ \textbf{to} $T_1$}{
update $\{\boldsymbol{\mathcal{W}}_i\}$ via  Eq. \ref{eqn:w_update};\\
update $\{\boldsymbol{\mathcal{Z}}_i\}$ via  Eq. \ref{eqn:z_update} and Alg. \ref{algorithm1};\\
$\boldsymbol{\mathcal{M}}_i \gets \boldsymbol{\mathcal{M}}_i + \boldsymbol{\mathcal{W}}_i - \boldsymbol{\mathcal{Z}}_i;$
}
$\{\boldsymbol{\mathcal{U}}^1_i, \boldsymbol{\mathcal{U}}^2_i, \boldsymbol{\mathcal{G}}_i\} \gets \texttt{decompose}(\{\boldsymbol{\mathcal{W}}_i\})$ via Alg. \ref{algorithm1};\\
\For ({~~~~~\textcolor{brown}{\textit{$\triangleright$ fine-tune}}}) {$t=1$ \textbf{to} $T_2$}{ 
$\boldsymbol{\mathcal{X}}, \boldsymbol{y} \gets \texttt{sample\_batch}(\boldsymbol{\mathcal{D}})$;\\
$\boldsymbol{\mathcal{X}}_{adv} \gets \texttt{PGD}(\boldsymbol{\mathcal{X}}, \boldsymbol{y})$ via Eq. \ref{eqn:pgd};\\
$\hat{\boldsymbol{y}} \gets \texttt{conv}(\{\boldsymbol{\mathcal{U}}^1_i, \boldsymbol{\mathcal{U}}^2_i, \boldsymbol{\mathcal{G}}_i\}, \boldsymbol{\mathcal{X}}_{adv})$ via Eq. \ref{eqn:convolution};\\
loss $\gets$ \texttt{cross\_entropy}($\hat{\boldsymbol{y}}, \boldsymbol{y}$);\\
\texttt{update}($\{\boldsymbol{\mathcal{U}}^1_i, \boldsymbol{\mathcal{U}}^2_i, \boldsymbol{\mathcal{G}}_i\}$, loss);\\
}
\caption{The overall procedure of CSTAR}
\label{algorithm2}
\end{algorithm}

\begin{table*}[ht]
\setlength{\tabcolsep}{10pt}
\centering
\scalebox{0.8}{
\begin{tabular}{
>{\columncolor[HTML]{FFFFFF}}l 
>{\columncolor[HTML]{FFFFFF}}c
>{\columncolor[HTML]{EFEFEF}}c 
>{\columncolor[HTML]{FFFFFF}}c
>{\columncolor[HTML]{FFFFFF}}c 
>{\columncolor[HTML]{EFEFEF}}c
>{\columncolor[HTML]{FFFFFF}}c 
>{\columncolor[HTML]{FFFFFF}}c
>{\columncolor[HTML]{EFEFEF}}c 
}
\toprule
\textbf{A.T.} & \multicolumn{2}{c}{\textbf{Friendly AT \cite{zhang2020attacks}}} & & \multicolumn{2}{c}{\textbf{TRADES \cite{zhang2019theoretically}}} & & \multicolumn{2}{c}{\textbf{MART \cite{wang2019improving}}} \\ \cmidrule{2-3} \cmidrule{5-6} \cmidrule{8-9}
C.R. & \textbf{RAP} & \textbf{CSTAR} &  & \textbf{RAP} & \textbf{CSTAR} &  & \textbf{RAP} & \textbf{CSTAR} \\ \midrule
& \multicolumn{2}{c}{\textbf{Pre-train 85.86 / 38.24}} &  & \multicolumn{2}{c}{\textbf{Pre-train 78.45 / 47.43}} &  & \multicolumn{2}{c}{\textbf{Pre-train 73.97 / 45.61}} \\
$2\times$ & 84.94 / 39.34 & \textbf{85.94 / 40.14} & \textbf{} & 78.47 / \textbf{46.92} & \textbf{80.00} / 44.81 & \multicolumn{1}{l}{} & 75.40 / 44.49 & \textbf{76.02 / 47.98} \\
$4\times$ & 83.24 / 37.10 & \textbf{85.86 / 40.29} & \textbf{} & 75.13 / 44.27 & \textbf{80.20 / 44.91} & \textbf{} & 71.23 / 44.60 & \textbf{77.08 / 44.87} \\
$8\times$ & 79.06 / 33.89 & \textbf{85.49 / 39.68} & \textbf{} & 69.82 / 39.58 & \textbf{79.48 / 44.94} & \textbf{} & 65.43 / 43.04 & \textbf{76.36 / 44.99} \\
$16\times$ & 74.06 / 29.01 & \textbf{85.02 / 38.70} & \textbf{} & 65.72 / 36.00 & \textbf{78.99 / 44.38} & \textbf{} & 57.74 / 38.57 & \textbf{74.57 / 46.25} \\
$32\times$ & 63.41 / 22.62 & \textbf{83.57 / 37.72} & \textbf{} & 56.47 / 30.75 & \textbf{77.49 / 42.59} & \textbf{} & 50.00 / 35.58 & \textbf{71.94 / 46.72} \\
$64\times$ & 54.30 / 20.59 & \textbf{81.32 / 35.44} & \textbf{} & 49.76 / 29.67 & \textbf{73.11 / 42.44} & \textbf{} & 42.40 / 31.76 & \textbf{66.33 / 44.95} \\
\bottomrule
\end{tabular}}
\caption{Performance of VGG-16 on CIFAR-10 using different adversarial training approaches. Here C.R. means compression ratio, and A.T. means adversarial training.}
\label{table:multiple_adv_training}
\end{table*}

\begin{table*}[t]
\setlength{\tabcolsep}{8pt}
\centering
\scalebox{0.8}{
\begin{tabular}{
>{\columncolor[HTML]{FFFFFF}}l
>{\columncolor[HTML]{FFFFFF}}c
>{\columncolor[HTML]{FFFFFF}}c 
>{\columncolor[HTML]{EFEFEF}}c
>{\columncolor[HTML]{FFFFFF}}c 
>{\columncolor[HTML]{EFEFEF}}c
>{\columncolor[HTML]{FFFFFF}}c 
>{\columncolor[HTML]{EFEFEF}}c
}
\toprule
 & & \multicolumn{6}{c}{\textbf{Methods / Compression Ratios}} \\ \cmidrule{3-8}
\textbf{Model Arch.} & \textbf{Pre-Train} & \textbf{\begin{tabular}[c]{@{}c@{}}RAP\\ 8x\end{tabular}} & \textbf{\begin{tabular}[c]{@{}c@{}}CSTAR\\ 8x\end{tabular}} & \textbf{\begin{tabular}[c]{@{}c@{}}RAP\\ 16x\end{tabular}} & \textbf{\begin{tabular}[c]{@{}c@{}}CSTAR\\ 16x\end{tabular}} & \textbf{\begin{tabular}[c]{@{}c@{}}RAP\\ 32x\end{tabular}} & \textbf{\begin{tabular}[c]{@{}c@{}}CSTAR\\ 32x\end{tabular}} \\ \midrule
DenseNet-121 & 80.91 / 45.36 & 71.74  43.66 & \textbf{78.87 / 48.07} & 57.42 / 36.23 & \textbf{74.91 / 45.38} & 36.79 / 26.33 & \textbf{63.12 / 37.71} \\
GoogleNet & 81.85 / 48.18 & 73.28 / 42.74 & \textbf{79.59 / 47.34} & 62.03 / 36.67 & \textbf{76.16 / 44.13} & 49.27 / 30.49 & \textbf{62.06 / 35.29} \\
VGG-19 & 74.64 / 43.33 & 65.94 / 40.79 & \textbf{73.60 / 45.95} & 57.35 / 35.95 & \textbf{73.88 / 45.87} & 47.45 / 31.73 & \textbf{72.69 / 45.59} \\
ResNet-34 & 82.64 / 47.85 & 79.05 / 47.34 & \textbf{83.52 / 50.35} & 73.93 / 44.62 & \textbf{81.95 / 48.69} & 66.45 / 40.13 & \textbf{79.63 / 48.24} \\
MobileNetV2 & 84.62 / 45.15 & 62.98 / 38.56 & \textbf{83.03 / 41.54} & 50.70 / 33.19 & \textbf{66.30 / 41.72} & 35.96 / 26.42 & \textbf{63.88 / 40.18} \\
\bottomrule
\end{tabular}}
\caption{Performance of benign / PGD-50 robust accuracy (\%) on CIFAR-10 using a wide range of model architectures with Standard AT. Here C.R. means compression ratio.}
\label{table:model_arch}
\end{table*}

\begin{table*}[ht]
\setlength{\tabcolsep}{10pt}
\centering
\scalebox{0.8}{
\begin{tabular}{
>{\columncolor[HTML]{FFFFFF}}l
>{\columncolor[HTML]{FFFFFF}}c
>{\columncolor[HTML]{EFEFEF}}c 
>{\columncolor[HTML]{FFFFFF}}c
>{\columncolor[HTML]{FFFFFF}}c 
>{\columncolor[HTML]{FFFFFF}}c
>{\columncolor[HTML]{EFEFEF}}c 
>{\columncolor[HTML]{FFFFFF}}c
}
\toprule
 & \multicolumn{3}{c}{\textbf{Top-1 Accuracy}} & \textbf{} & \multicolumn{3}{c}{\textbf{Top-5 Accuracy}} \\ \cmidrule{2-4} \cmidrule{6-8}
C.R. & \textbf{RAP} & \textbf{CSTAR} & $\boldsymbol{\Delta}$ & \textbf{} & \textbf{RAP} & \textbf{CSTAR} & $\boldsymbol{\Delta}$ \\ \midrule
 & \multicolumn{3}{c}{\textbf{Pre-train 50.75 / 26.57}} &  & \multicolumn{3}{c}{\textbf{Pre-train 75.25 / 51.56}}\\
$2\times$ & 43.40 / 22.22 & \textbf{51.36 / 25.75} & +7.96 / +3.53 &  & 68.69 / 45.16 & \textbf{75.28 / 51.42} & +6.59 / +6.26 \\
$4\times$ & 35.29 / 17.58 & \textbf{41.11 / 21.05} & +5.82 / +3.47 &  & 60.32 / 37.71 & \textbf{66.60 / 43.28} & +6.28 / +5.57 \\
$6\times$ & 29.40 / 14.50 & \textbf{34.92 / 17.35} & +5.52 / +2.85 &  & 53.06 / 31.95 & \textbf{59.92 / 37.41} & +6.86 / +5.46 \\
$8\times$ & 23.24 / 11.72 & \textbf{30.29 / 14.84} & +7.05 / +3.12 &  & 45.31 / 26.88 & \textbf{54.60 / 33.10} & +9.29 / +6.22 \\
$10\times$ & 19.09 /  9.72 & \textbf{28.91 / 14.10} & +9.82 / +4.38 &  & 39.41 / 23.14 & \textbf{52.82 / 31.86} & +13.41 / +8.72 \\
$16\times$ & 9.34 / 5.05 & \textbf{17.92 / 9.32} & +8.58 / +4.27 &  & 22.93 / 13.20 & \textbf{37.83 / 22.22} & +14.90 / +9.02 \\
\bottomrule
\end{tabular}}
\caption{Performance of benign / robust accuracy (\%) for ResNet-18 on ImageNet dataset using Fast AT \cite{wong2020fast}. Here C.R. means compression ratio.}
\label{table:imagenet_rn18}
\end{table*}

\ul{Analysis.} Because 1) the rank values directly determine the model capacity, computational cost, and memory footprint; and 2) different layers exhibit different low-rankness characteristics, a proper rank selection scheme is very critical and important to achieve a high-performance low-rank DNN model. Unfortunately, as analyzed in Section \ref{sec:related_works}, to date the rank selection for the tensor decomposed models are still performed in a costly and time-consuming way. In particular, consider each low-rank layer in modern DNNs may have multiple to-be-set tensor ranks, e.g., Tucker-2 format needs to configure two rank values per layer, the existing rank selection scheme will become even more expensive.

\ul{Our Proposed Solution.} To overcome this challenge, we propose a low-cost automatic rank selection scheme for low-rank adversarial robust DNN models. Our key idea is to \ul{leverage the singular value information to globally select the suitable tensor rank for each individual layer.} To be specific, considering mathematically the magnitude of each singular value strongly represents the importance of the corresponding rank component, e.g., with the sorted singular values $\sigma^1_i$ for $\boldsymbol{\mathcal{U}}^1$, $\boldsymbol{\mathcal{U}}^1_k = \sigma^{1}_{1} u^{1}_{1} v^{1T}_{1} +...+\sigma^{1}_{k} u^{1}_{k} v^{1T}_{k}$ is the best rank-$k$ approximation for $\boldsymbol{\mathcal{U}}^{1}$. Hence we can sort all the singular values from the entire model and select the largest ones and their corresponding rank components under the target compression budget. As illustrated in Fig. \ref{fig:global_rank}, this global comparison and selection strategy can determine the suitable tensor rank values for different layers in an automatic way, thereby significantly facilitating the overall compression workflow. Noticed that here the desired singular values have already been available after low-rank projection (see Line 6 and 8 in Alg. \ref{algorithm1}), so it is naturally integrated to the entire compression procedure with very low computing overhead. Also, as showed in Fig. \ref{fig:global_rank}, such a sorting-and-selection process can be performed in each epoch to dynamically adjust the rank selection as training progresses.

\textbf{Fine-tuning.} Upon the iterative update finishes, the model has been gradually imposed on the desired low-tensor-rank property, and then we can decompose it with the automatically selected ranks and fine-tune it in the low-rank format. Algorithm \ref{algorithm2} summarizes the CSTAR procedure.

\section{Experiments}
\label{sec:experiments}

To demonstrate the effectiveness of CSTAR, we evaluate various model architectures with multiple adversarial training objectives on different datasets. To be specific, we evaluate the benign and robust accuracy of VGG-16/19, ResNet-18/34, DenseNet-121, GoogleNet and MobileNet-V2. In addition, several adversarial training approaches, including the Standard PGD \cite{madry2017towards}, Friendly AT \cite{zhang2020attacks}, TRADES \cite{zhang2019theoretically}, MART \cite{wang2019improving}, and Fast AT \cite{wong2020fast} are adopted to show the generality of our approach. The models are compressed on CIFAR-10/100 and ImageNet datasets with different compression ratios ranging from $2\times$ to $64\times$. On CIFAR-10/100 $L_\infty$ is selected with $\Delta = 8/255$, and PGD with step size $\alpha = 2/255$ serves to generate adversarial examples. Here the number of PGD iterations for training and testing are 10 and 50, respectively. Other hyper-parameter settings can be found in the Appendix.



\textbf{Comparison with Robust Aware Pruning (RAP) on CIFAR-10/100.} Table \ref{table:sota} shows our comparison with the RAP approach \cite{ye2019adversarial}. 
For fair comparison, we perform compression on the same pre-train models with the same number of training epochs. 
In addition, we also report the performance of Least Weight Magnitude (LWM)-based approach \cite{sehwag2019towards} in this table. It is seen that our proposed approach consistently outperforms RAP with different compression ratios on both benign accuracy and robust accuracy. In particular, with $64\times$ compression, our CSTAR can achieve up to $20.07\%$ and $11.91\%$ improvement for benign accuracy and robust accuracy, respectively. For the results of CIFAR-100 dataset, see the Appendix.

\textbf{Comparison with HYDRA on CIFAR-10.} HYDRA \cite{sehwag2020hydra} is AN importance-score based adversarial pruning approach. Notice that because 1) HYDRA does not release code for their \textbf{structured pruning} version; and 2) HYDRA only reports very limited data point (VGG-16 on CIFAR-10) for its performance with structured pruning, in Table \ref{table:hydra} we only list the benign and robust accuracy of our approach with the C.Rs. (2$\times$ and 10$\times$) that HYDRA reports. From this table it is seen that our method shows very significant performance improvement over HYDRA. With 10 $\times$ C.R., our CSTAR approach can achieve $60.99\%$ of benign accuracy and $28.48\%$ robust accuracy increase.

\begin{table}[t]
\setlength{\tabcolsep}{12pt}
\centering
\scalebox{0.8}{
\begin{tabular}{
>{\columncolor[HTML]{FFFFFF}}l 
>{\columncolor[HTML]{FFFFFF}}c
>{\columncolor[HTML]{EFEFEF}}c 
>{\columncolor[HTML]{FFFFFF}}c
}
\toprule
C.R. & \textbf{HYDRA} & \textbf{CSTAR} & $\boldsymbol{\Delta}$  \\
\midrule
$2\times$ & 52.90 / 38.00 & \textbf{80.00 / 44.81} & +27.10 / +6.81 \\
$10\times$ & 18.30 / 16.70 & \textbf{79.29 / 45.18} & +60.99 / +28.48 \\
\bottomrule
\end{tabular}}
\caption{HYDRA vs CSTAR on CIFAR-10 using TRADES. Here C.R. means compression ratio.}
\label{table:hydra}
\end{table}

\textbf{Stronger Baseline: RAP or HYDRA?} Comparing Table \ref{table:sota} and Table \ref{table:hydra}, it is seen that with the same (2$\times$) or even higher (16$\times$) compression ratio, RAP shows much higher benign and robust accuracy than HYDRA on the same pre-trained VGG-16 model. Such phenomenon shows that the least weight magnitude-based structured pruning seems more suitable for obtaining adversarial robustness than the importance score-based structured pruning. In addition, considering HYDRA only reports its \textbf{structured pruning} performance for VGG-16 on CIFAR-10; therefore, in our following experiments involved with other models and dataset, we select RAP (Column pruning) as the baseline.


\textbf{Generalization Across Different Adversarial Training Approaches.} To demonstrate the generality of CSTAR, we evaluate its performance for compressing VGG-16 model via using other three popular adversarial training methods (FAT, TRADES and MART). As shown in Table \ref{table:multiple_adv_training}, our solution can consistently outperform RAP method with different compression ratios on both benign and robust accuracy.

\textbf{Generalization Across Different Model Architectures.} We also evaluate the generality of CSTAR on different CNN architectures. As shown in Table \ref{table:model_arch}, our CSTAR solution achieves both higher benign and robust accuracy than RAP across multiple models sizes and architectures.


\begin{table}[t]
\centering
\scalebox{0.75}{
\begin{tabular}{
>{\columncolor[HTML]{FFFFFF}}l
>{\columncolor[HTML]{FFFFFF}}c
>{\columncolor[HTML]{EFEFEF}}c 
>{\columncolor[HTML]{FFFFFF}}c
>{\columncolor[HTML]{FFFFFF}}c 
>{\columncolor[HTML]{EFEFEF}}c
}
\toprule
\multirow{2}{*}{\textbf{C.R.}} & \multicolumn{2}{c}{\textbf{VGG-16 / CPU}} & & \multicolumn{2}{c}{\textbf{ResNet-18 / GPU}} \\ \cmidrule{2-3} \cmidrule{5-6}
 & \textbf{RAP} & \textbf{CSTAR} & & \textbf{RAP} & \textbf{CSTAR} \\ \midrule
P.T. & 9.98ms & 9.98ms & & 0.0801ms & 0.0801ms\\
2$\times$ & 9.98ms & 6.75ms/$\downarrow$1.48$\times$ & & 0.0801ms & 0.0562ms/$\downarrow$1.43$\times$\\
4$\times$ & 9.98ms & 4.38ms/$\downarrow$2.28$\times$ & & 0.0801ms & 0.0437ms/$\downarrow$1.83$\times$\\
\bottomrule
\end{tabular}}
\caption{Inference time per image of executing CSTAR models for CIFAR-10 data compared with unstructured pruning.}

\label{table:speed_up}
\end{table}

\textbf{Performance on ImageNet Dataset.} Table \ref{table:imagenet_rn18} shows the performance of our approach for compressing ResNet-18 on ImageNet dataset. It is seen that in this challenging task CSTAR achieves consistently better performance than RAP. For instance, CSTAR can obtain $8.58\%$ benign accuracy gain and $4.27\%$ robust accuracy gain with $16 \times$ C.R..

\textbf{Runtime Speedup on CPU/GPU.} We also measure the speedup of executing CSTAR model on the off-the-shelf hardware. Table \ref{table:speed_up} reports the inference time with batch size = 1 for the original and compressed models on both CPU (AMD Ryzen 9 5900HX) and GPU (NVIDIA GeForce RTX 3090). It is seen that the low-rank structuredness of CSTAR indeed brings practical speedup.

\section{Conclusion}

This paper proposes CSTAR, an efficient approach that 
can simultaneously impose high low-rankness-based compactness, high structuredness, and high adversarial robustness on the DNN models. By using a unified framework to gradually impose the low-tensor-rankness and automatically select the rank setting during the adversarial training procedure, CSTAR can make the compressed DNNs achieve high compression performance and strong adversarial robustness at the same time. 

\paragraph{Acknowledgments} This work was partially supported by National Science Foundation under Grant CCF-1955909, RTML, CMMI and CPS programs.

\newpage
\bibliography{aaai23.bib}

\end{document}


\maketitle

\section{Hyper-parameter Settings}
For the underlying training procedure of all the evaluation datasets, the optimizer is set as SGD with an initial learning rate of $0.1$, which is gradually reduced by a factor of $10$ at the $25\%$, $50\%$ and $75\%$ of the total training epochs. For CIFAR-10/100, we adopt the setting used in \cite{ye2019adversarial,sehwag2020hydra} for adversarial training. To be specific, $L_\infty$ is selected with $\Delta = 8/255$, and PGD with step size $\alpha = 2/255$ serves to generate adversarial examples. Here the number of PGD iterations for training and test are 10 and 50, respectively. In addition, we use 100 training epochs for the low-rank regularization phase and fine-tuning phase. For ImageNet, considering the very heavy computing demand when performing PGD adversarial training on large-scale ImageNet dataset, similar to other works, e.g., \cite{fu2021double}, we choose to perform Fast Adversarial Training \cite{wong2020fast} with $\Delta = 4/255$ and $\alpha = 5/255$. Here the number of iterations is set as 1 and 10 for training and testing, respectively. Also, the number of training epochs for both the low-rank regularization phase and fine-tuning phase is set at 20.

\section{Performance on CIFAR-100 Dataset.}
 Table \ref{table:cifar100} reports the accuracy performance when compressing VGG-16 and ResNet-18 on CIFAR-100 dataset. It is seen that our approach achieves higher benign and robust accuracy than RAP with different compression ratios. For instance, when compressing VGG-16 network to 32$\times$ smaller model size, CSTAR enjoys $16.52\%$ higher benign accuracy and $7.40\%$ higher robust accuracy.

\begin{table}[t]
\setlength{\tabcolsep}{6pt}
\centering
\caption{Performance of VGG-16 on CIFAR-10 with/without using low-rank regularization using Standard AT. Here C.R. means compression ratio.}
\scalebox{1.0}{
\begin{tabular}{
>{\columncolor[HTML]{FFFFFF}}l 
>{\columncolor[HTML]{FFFFFF}}c 
>{\columncolor[HTML]{EFEFEF}}c 
>{\columncolor[HTML]{FFFFFF}}c
}
\toprule
C.R. & \textbf{Tucker} & \textbf{CSTAR} & \textbf{$\boldsymbol{\Delta}$} \\ \midrule
$2\times$ & 69.09 / 42.47 & \textbf{79.71 / 47.23} & +10.62 / +4.76 \\
$4\times$ & 67.22 / 42.32 & \textbf{79.64 / 47.06} & +12.42 / +4.74 \\
$8\times$ & 59.94 / 38.21 & \textbf{79.23 / 46.42} & +19.29 / +8.21 \\
$16\times$ & 56.57 / 36.02 & \textbf{77.82 / 46.13} & +21.25 / +10.11 \\
$32\times$ & 50.87 / 33.91 & \textbf{75.82 / 45.43} & +24.95 / +11.52 \\
$64\times$ & 49.92 / 33.24 & \textbf{70.77 / 43.13} & +20.85 / +9.89 \\
\bottomrule
\end{tabular}}
\label{table:ablation_simple_tucker}
\end{table}

\begin{table*}[h]
\setlength{\tabcolsep}{10pt}
\centering
\scalebox{0.95}{
\begin{tabular}{
>{\columncolor[HTML]{FFFFFF}}l
>{\columncolor[HTML]{FFFFFF}}c
>{\columncolor[HTML]{EFEFEF}}c 
>{\columncolor[HTML]{FFFFFF}}c
>{\columncolor[HTML]{FFFFFF}}c 
>{\columncolor[HTML]{FFFFFF}}c
>{\columncolor[HTML]{EFEFEF}}c 
>{\columncolor[HTML]{FFFFFF}}c
}
\toprule
 & \multicolumn{3}{c}{\textbf{VGG-16}} &  & \multicolumn{3}{c}{\textbf{ResNet-18}} \\
\cmidrule{2-4} \cmidrule{6-8}
C.R. & \textbf{RAP} & \textbf{CSTAR} & $\boldsymbol{\Delta}$ &  & \textbf{RAP} & \textbf{CSTAR} & $\boldsymbol{\Delta}$ \\
\midrule
& \multicolumn{3}{c}{\textbf{Pre-train 48.99 / 22.44}} &  & \multicolumn{3}{c}{\textbf{Pre-train 57.73 / 27.29}}\\
$2\times$ & 49.22 / 22.72 & \textbf{50.51 / 24.33} & +1.29 / +1.61 &  & 57.26 / 26.09 & \textbf{57.69 / 26.69} & +0.43 / +0.60 \\
$4\times$ & 47.44 / 22.07 & \textbf{50.02 / 24.12} & +2.58 / +2.05 &  & 55.74 / \textbf{26.13} & \textbf{56.42} / 25.76 & +0.68 / -0.37 \\
$8\times$ & 44.50 / 18.82 & \textbf{47.55 / 23.22} & +3.05 / +4.40 &  & 53.38 / 24.52 & \textbf{54.73 / 25.54} & +1.35 / +1.02 \\
$16\times$ & 38.87 / 18.27 & \textbf{47.91 / 23.25} & +9.04 / +4.98 &  & 47.04 / 23.10 & \textbf{51.88 / 27.47} & +4.84 / +4.37 \\
$32\times$ & 29.42 / 15.46 & \textbf{45.94 / 22.86} & +16.52 / +7.40 &  & 35.27 / 18.84 & \textbf{47.69 / 23.46} & +12.42 / +4.62 \\
$64\times$ & 23.20 / 13.00 & \textbf{36.21 / 19.62} & +13.01 / +6.62 &  & 25.02 / 13.95 & \textbf{38.67 / 20.82} & +13.65 / +6.87 \\
\bottomrule
\end{tabular}}
\caption{Performance of benign / PGD-50 robust accuracy (\%) on CIFAR-100 dataset using Standard AT. Here C.R. means compression ratio.}
\label{table:cifar100}
\end{table*}

\section{Performance against Black-box Attack.}

To show the robustness against black-box attacks, we generate adversarial examples using a wide range of full size models and then launch the adversarial attack against our CSTAR models. As reported in Table \ref{table:blackbox}, our compact models show high robustness against the black-box adversarial attack prepared by different types of model architecture. Compared with RAP approach, our CSTAR demonstrates much stronger black-box adversarial robustness.

\begin{table*}[h]
\centering
\scalebox{0.9}{
\begin{tabular}{
>{\columncolor[HTML]{FFFFFF}}l 
>{\columncolor[HTML]{FFFFFF}}c
>{\columncolor[HTML]{EFEFEF}}c 
>{\columncolor[HTML]{FFFFFF}}c
>{\columncolor[HTML]{EFEFEF}}c 
>{\columncolor[HTML]{FFFFFF}}c
>{\columncolor[HTML]{EFEFEF}}c 
>{\columncolor[HTML]{FFFFFF}}c
>{\columncolor[HTML]{EFEFEF}}c 
>{\columncolor[HTML]{FFFFFF}}c
>{\columncolor[HTML]{EFEFEF}}c 
>{\columncolor[HTML]{FFFFFF}}c
>{\columncolor[HTML]{EFEFEF}}c 
}
\toprule
\multicolumn{1}{l}{} & \multicolumn{12}{c}{\textbf{Evaluation using Method / Model Architecture / Compression Ratio}} \\ \cmidrule{2-13}
\multicolumn{1}{l}{\textbf{}} & \multicolumn{1}{l}{\textbf{RAP}} & \multicolumn{1}{l}{\cellcolor[HTML]{EFEFEF}\textbf{CSTAR}} & \multicolumn{1}{l}{\textbf{RAP}} & \multicolumn{1}{l}{\cellcolor[HTML]{EFEFEF}\textbf{CSTAR}} & \multicolumn{1}{l}{\textbf{RAP}} & \multicolumn{1}{l}{\cellcolor[HTML]{EFEFEF}\textbf{CSTAR}} & \multicolumn{1}{l}{\textbf{RAP}} & \multicolumn{1}{l}{\cellcolor[HTML]{EFEFEF}\textbf{CSTAR}} & \multicolumn{1}{l}{\textbf{RAP}} & \multicolumn{1}{l}{\cellcolor[HTML]{EFEFEF}\textbf{CSTAR}} & \multicolumn{1}{l}{\textbf{RAP}} & \multicolumn{1}{l}{\cellcolor[HTML]{EFEFEF}\textbf{CSTAR}} \\
\textbf{\begin{tabular}[c]{@{}c@{}}Black-Box\\ to Gen. Adv.\end{tabular}} & \multicolumn{2}{c}{\begin{tabular}[c]{@{}c@{}}VGG-16 \\ 8$\times$\end{tabular}} & \multicolumn{2}{c}{\begin{tabular}[c]{@{}c@{}}VGG-16 \\ 32$\times$\end{tabular}} & \multicolumn{2}{c}{\begin{tabular}[c]{@{}c@{}}ResNet-18\\ 8$\times$\end{tabular}} & \multicolumn{2}{c}{\begin{tabular}[c]{@{}c@{}}ResNet-18\\ 32$\times$\end{tabular}} & \multicolumn{2}{c}{\begin{tabular}[c]{@{}c@{}}DenseNet-121\\ 8$\times$\end{tabular}} & \multicolumn{2}{c}{\begin{tabular}[c]{@{}c@{}}DenseNet-121\\ 32$\times$\end{tabular}} \\ \midrule
\textbf{VGG-16} & 55.17 & \textbf{58.60} & 46.82 & \textbf{57.94} & 58.99 & \textbf{63.12} & 51.76 & \textbf{59.71} & 56.56 & \textbf{61.48} & 34.31 & \textbf{52.11} \\
\textbf{ResNet-18} & 55.72 & \textbf{58.97} & 49.97 & \textbf{58.50} & 57.71 & \textbf{58.85} & 52.21 & \textbf{57.41} & 57.40 & \textbf{59.71} & 34.32 & \textbf{52.75} \\
\textbf{DenseNet121} & 55.47 & \textbf{61.11} & 46.86 & \textbf{49.44} & 58.59 & \textbf{62.35} & 51.72 & \textbf{59.23} & 55.95 & \textbf{58.74} & 34.08 & \textbf{51.62} \\
\textbf{VGG-19} & 54.14 & \textbf{60.80} & 45.98 & \textbf{58.28} & 59.29 & \textbf{64.69} & 50.62 & \textbf{60.24} & 57.22 & \textbf{62.03} & 33.75 & \textbf{51.18} \\
\textbf{ResNet-34} & 56.56 & \textbf{60.06} & 47.14 & \textbf{58.72} & 58.60 & \textbf{61.15} & 52.79 & \textbf{58.75} & 58.24 & \textbf{60.46} & 34.66 & \textbf{52.72} \\
\textbf{GoogleNet} & 55.34 & \textbf{59.59} & 46.43 & \textbf{57.92} & 58.02 & \textbf{61.62} & 51.02 & \textbf{58.58} & 56.96 & \textbf{59.40} & 34.10 & \textbf{51.18} \\
\textbf{MobileNet} & 55.41 & \textbf{59.40} & 46.78 & \textbf{58.26} & 57.27 & \textbf{60.43} & 51.65 & \textbf{57.71} & 62.84 & \textbf{65.90} & 35.54 & \textbf{56.99} \\
\bottomrule
\end{tabular}
}
\caption{Robust Accuracy ($\%$) against black-box attacks on CIFAR-10.}
\label{table:blackbox}
\end{table*}

\begin{table}[t]
\setlength{\tabcolsep}{5pt}
\centering
\scalebox{0.9}{
\begin{tabular}{lccc}
\toprule
\textbf{Scheme} / C.R. & $4\times$ & $16\times$ & $64\times$ \\\midrule
\textbf{Uniform} & 78.51 / 46.36 & 73.67 / 44.60 & 62.66 / 39.14 \\
\textbf{Global} & \textbf{79.64} / \textbf{47.06} & \textbf{77.82} / \textbf{46.13} & \textbf{70.77 / 43.13} \\
\textbf{Global MM} & 79.53 / 47.03 & 76.45 /46.08 & 68.96 / 42.64 \\
\textbf{Global Std} & 79.34 / 46.81 & 76.32 / 46.03 & 60.21 / 38.32 \\
\bottomrule
\end{tabular}
}
\caption{Performance of VGG-16 on CIFAR-10 with different rank selection schemes. Here C.R. means compression ratio.}
\label{table:ranks_scheme}
\end{table}

\section{Ablation Study on the Effect of Low-rank Regularization.}

We examine the effect of low-rank regularization process via evaluating the performance of adversarial training without using low-rank regularization. To be specific, in this ablation study we directly decompose the pre-train models into Tucker-2 format, and fine-tune it with adversarial training. As observed in Table \ref{table:ablation_simple_tucker},  our CSTAR enjoys significant performance improvement than the low-rank regularization-free method. For instance, our approach can achieve up to $24.95\%$ increase for benign accuracy and $11.52\%$ increase for robust accuracy with $32 \times$ compression ratio.

\section{Ablation Study the Effect of Automatic Rank Selection.} 

\paragraph{Effect of different rank selection schemes.} Considering the potential wide value range of $\sigma^1$ and $\sigma^2$ for different layers, we investigate the necessity for normalization of singular values.
Here two possible variants of our approach are explored: performing min-max (\textbf{``MM"}) normalization and standard (\textbf{``Std"}) normalization on $\sigma^1$ and $\sigma^2$ of each layer before putting them together to the global sorting. 
As shown in Table \ref{table:ranks_scheme}, the normalization-free approach (\textbf{``Global"}) consistently achieves better performance than its normalized variants and another baseline (\textbf{``Uniform"}) where each layer is assigned with the same compression ratio. Therefore, we suggest the normalization for singular values is not needed.

\begin{figure*}[h]
\centering
\includegraphics[width=0.9\textwidth]{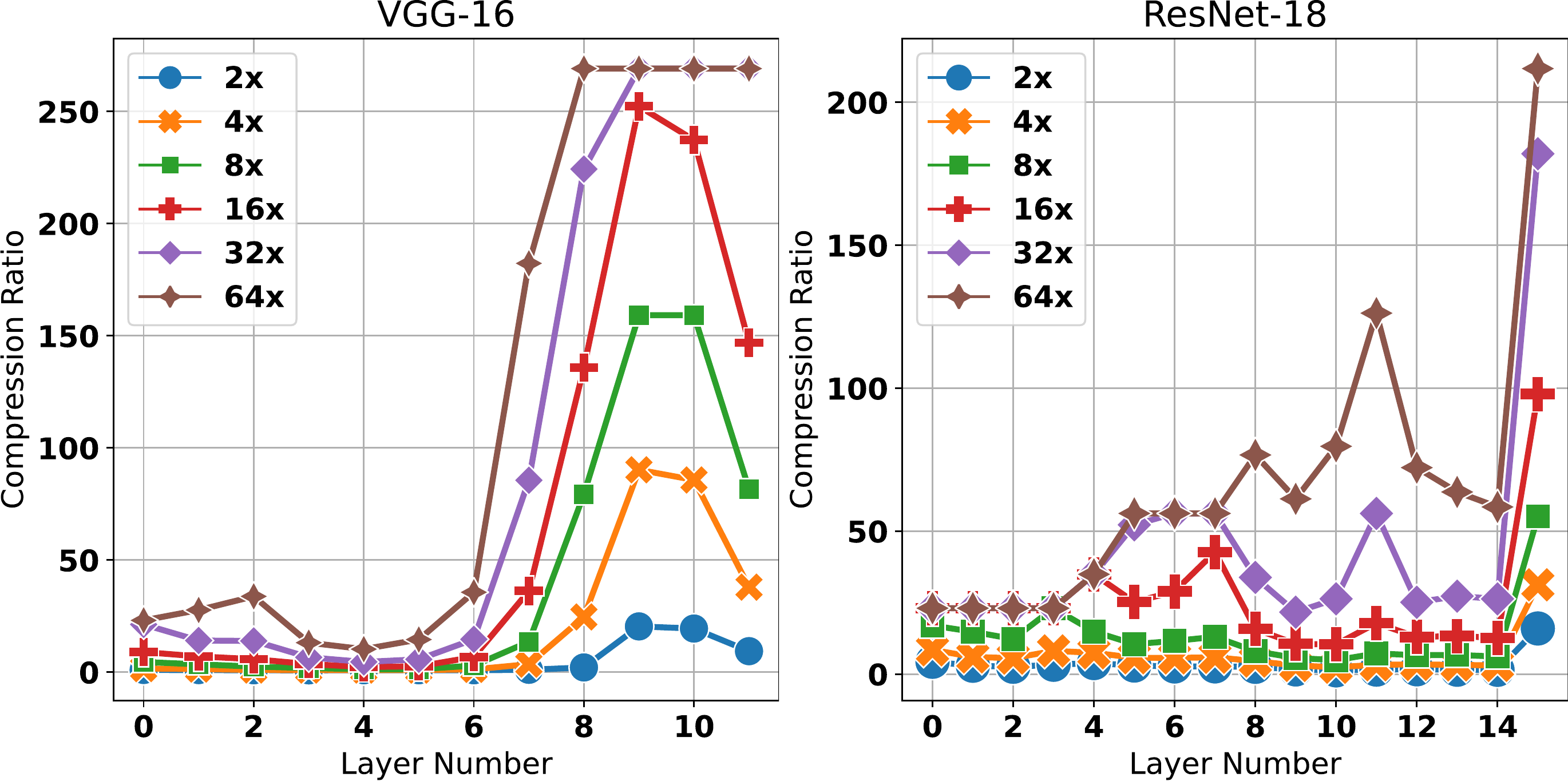}
\caption{The layer-wise compression ratio for VGG-16 and 
ResNet-18 on CIFAR-10 after automatic rank selection.}
\label{figure:layer_comp}
\end{figure*}

\paragraph{Details on layer-wise compression ratio and layer rank settings.}
For the layer-wise rank settings, we assert a minimum rank of $[8, 8]$ per layer to ensure the minimum layer complexity for stable training. Figure \ref{figure:layer_comp} illustrates the layer-wise compression ratio after performing the proposed automatic rank selection. We also report the specific rank settings with different compression ratios for the VGG-16 model in Table \ref{table:vgg16_ranks}, and for ResNet-18 in Table \ref{table:resnet18_ranks}. It is seen that different layers are compressed with different ratios as individual tensor rank setting is assigned to each of them. 

\begin{table*}[h]
\setlength{\tabcolsep}{10pt}
\centering
\caption{Layer Tucker-2 ranks and layer compression ratios of VGG-16 on CIFAR-10 after using automatic rank selection.}
\begin{tabular}{@{}lccccc@{}}
\toprule
\textbf{Layer Name} & \textbf{\begin{tabular}[c]{@{}c@{}}Weight\\ Shape\end{tabular}} & \textbf{\begin{tabular}[c]{@{}c@{}}Tucker-2\\ Ranks\end{tabular}} & \textbf{\begin{tabular}[c]{@{}c@{}}Uncomp.\\ Params\end{tabular}} & \textbf{\begin{tabular}[c]{@{}c@{}}Comp.\\ Params\end{tabular}} & \textbf{\begin{tabular}[c]{@{}c@{}}Layer\\ C.R.\end{tabular}} \\ \midrule
\addlinespace[0.5em]
\multicolumn{6}{c}{\textbf{8$\times$ Compression Ratio}} \\
features.3 & (64, 64, 3, 3) & {[}27, 21{]} & 36.9K & 8.18K & 4.5$\times$ \\
features.7 & (128, 64, 3, 3) & {[}45, 33{]} & 73.7K & 21.2K & 3.5$\times$ \\
features.10 & (128, 128, 3, 3) & {[}74, 63{]} & 147K & 59.5K & 2.5$\times$ \\
features.14 & (256, 128, 3, 3) & {[}132, 97{]} & 295K & 161K & 1.8$\times$ \\
features.17 & (256, 256, 3, 3) & {[}196, 203{]} & 590K & 460K & 1.3$\times$ \\
features.20 & (256, 256, 3, 3) & {[}197, 193{]} & 590K & 442K & 1.3$\times$ \\
features.24 & (512, 256, 3, 3) & {[}177, 170{]} & 1.18M & 405K & 2.9$\times$ \\
features.27 & (512, 512, 3, 3) & {[}93, 94{]} & 2.36M & 174K & 13.5$\times$ \\
features.30 & (512, 512, 3, 3) & {[}24, 24{]} & 2.36M & 29.8K & 79.3$\times$ \\
features.34 & (512, 512, 3, 3) & {[}12, 14{]} & 2.36M & 14.8K & 159.2$\times$ \\
features.37 & (512, 512, 3, 3) & {[}14, 12{]} & 2.36M & 14.8K & 159.2$\times$ \\
features.40 & (512, 512, 3, 3) & {[}27, 20{]} & 2.36M & 28.9K & 81.6$\times$ \\

\addlinespace[0.5em]
\multicolumn{6}{c}{\textbf{16$\times$ Compression Ratio}} \\

features.3 & (64, 64, 3, 3) & {[}17, 14{]} & 36.9K & 4.13K & 8.9$\times$ \\
features.7 & (128, 64, 3, 3) & {[}28, 22{]} & 73.7K & 10.5K & 7.0$\times$ \\
features.10 & (128, 128, 3, 3) & {[}45, 37{]} & 147K & 25.5K & 5.8$\times$ \\
features.14 & (256, 128, 3, 3) & {[}87, 66{]} & 295K & 82.4K & 3.6$\times$ \\
features.17 & (256, 256, 3, 3) & {[}145, 142{]} & 590K & 259K & 2.3$\times$ \\
features.20 & (256, 256, 3, 3) & {[}132, 134{]} & 590K & 227K & 2.6$\times$ \\
features.24 & (512, 256, 3, 3) & {[}102, 104{]} & 1.18M & 174K & 6.8$\times$ \\
features.27 & (512, 512, 3, 3) & {[}47, 44{]} & 2.36M & 65.2K & 36.2$\times$ \\
features.30 & (512, 512, 3, 3) & {[}15, 15{]} & 2.36M & 17.4K & 135.7$\times$ \\
features.34 & (512, 512, 3, 3) & {[}8, 9{]} & 2.36M & 9.35K & 252.3$\times$ \\
features.37 & (512, 512, 3, 3) & {[}9, 9{]} & 2.36M & 9.94K & 237.2$\times$ \\
features.40 & (512, 512, 3, 3) & {[}16, 12{]} & 2.36M & 16.1K & 146.9$\times$ \\

\addlinespace[0.5em]
\multicolumn{6}{c}{\textbf{32$\times$ Compression Ratio}} \\

features.3 & (64, 64, 3, 3) & {[}9, 8{]} & 36.9K & 1.74K & 21.2$\times$ \\
features.7 & (128, 64, 3, 3) & {[}18, 13{]} & 73.7K & 5.24K & 14.1$\times$ \\
features.10 & (128, 128, 3, 3) & {[}25, 21{]} & 147K & 10.6K & 13.9$\times$ \\
features.14 & (256, 128, 3, 3) & {[}56, 48{]} & 295K & 44.7K & 6.6$\times$ \\
features.17 & (256, 256, 3, 3) & {[}97, 93{]} & 590K & 130K & 4.5$\times$ \\
features.20 & (256, 256, 3, 3) & {[}81, 84{]} & 590K & 103K & 5.7$\times$ \\
features.24 & (512, 256, 3, 3) & {[}59, 64{]} & 1.18M & 80.6K & 14.6$\times$ \\
features.27 & (512, 512, 3, 3) & {[}24, 21{]} & 2.36M & 27.6K & 85.6$\times$ \\
features.30 & (512, 512, 3, 3) & {[}8, 11{]} & 2.36M & 10.5K & 224.3$\times$ \\
features.34 & (512, 512, 3, 3) & {[}8, 8{]} & 2.36M & 8.77K & 269.1$\times$ \\
features.37 & (512, 512, 3, 3) & {[}8, 8{]} & 2.36M & 8.77K & 269.1$\times$ \\
features.40 & (512, 512, 3, 3) & {[}8, 8{]} & 2.36M & 8.77K & 269.1$\times$ \\
\bottomrule
\end{tabular}
\label{table:vgg16_ranks}
\end{table*}

\begin{table*}[h]
\setlength{\tabcolsep}{10pt}
\centering
\caption{Layer Tucker-2 ranks and layer compression ratios of ResNet-18 on CIFAR-10 with after using automatic rank selection.}
\scalebox{0.87}{
\begin{tabular}{@{}lccccc@{}}
\toprule
\textbf{Layer Name} & \textbf{\begin{tabular}[c]{@{}c@{}}Weight\\ Shape\end{tabular}} & \textbf{\begin{tabular}[c]{@{}c@{}}Tucker-2\\ Ranks\end{tabular}} & \textbf{\begin{tabular}[c]{@{}c@{}}Uncomp.\\ Params\end{tabular}} & \textbf{\begin{tabular}[c]{@{}c@{}}Comp.\\ Params\end{tabular}} & \textbf{\begin{tabular}[c]{@{}c@{}}Layer\\ C.R.\end{tabular}} \\ \midrule
\multicolumn{6}{c}{\textbf{8$\times$ Compression Ratio}} \\
layer1.0.conv1 & (64, 64, 3, 3) & {[}11, 9{]} & 36.9K & 2.17K & 17.0$\times$ \\
layer1.0.conv2 & (64, 64, 3, 3) & {[}12, 10{]} & 36.9K & 2.49K & 14.8$\times$ \\
layer1.1.conv1 & (64, 64, 3, 3) & {[}13, 12{]} & 36.9K & 3K & 12.3$\times$ \\
layer1.1.conv2 & (64, 64, 3, 3) & {[}8, 8{]} & 36.9K & 1.6K & 23.0$\times$ \\
layer2.0.conv1 & (128, 64, 3, 3) & {[}16, 14{]} & 73.7K & 4.96K & 14.9$\times$ \\
layer2.0.conv2 & (128, 128, 3, 3) & {[}32, 24{]} & 147K & 14.1K & 10.5$\times$ \\
layer2.1.conv1 & (128, 128, 3, 3) & {[}28, 24{]} & 147K & 12.7K & 11.6$\times$ \\
layer2.1.conv2 & (128, 128, 3, 3) & {[}25, 23{]} & 147K & 11.3K & 13.0$\times$ \\
layer3.0.conv1 & (256, 128, 3, 3) & {[}48, 44{]} & 295K & 36.9K & 8.0$\times$ \\
layer3.0.conv2 & (256, 256, 3, 3) & {[}94, 73{]} & 590K & 105K & 5.6$\times$ \\
layer3.1.conv1 & (256, 256, 3, 3) & {[}95, 87{]} & 590K & 121K & 4.9$\times$ \\
layer3.1.conv2 & (256, 256, 3, 3) & {[}70, 72{]} & 590K & 81.7K & 7.2$\times$ \\
layer4.0.conv1 & (512, 256, 3, 3) & {[}104, 105{]} & 1.18M & 178K & 6.6$\times$ \\
layer4.0.conv2 & (512, 512, 3, 3) & {[}159, 139{]} & 2.36M & 351K & 6.7$\times$ \\
layer4.1.conv1 & (512, 512, 3, 3) & {[}166, 150{]} & 2.36M & 386K & 6.1$\times$ \\
layer4.1.conv2 & (512, 512, 3, 3) & {[}32, 33{]} & 2.36M & 42.8K & 55.1$\times$ \\

\multicolumn{6}{c}{\textbf{16$\times$ Compression Ratio}} \\

layer1.0.conv1 & (64, 64, 3, 3) & {[}8, 8{]} & 36.9K & 1.6K & 23.0$\times$ \\
layer1.0.conv2 & (64, 64, 3, 3) & {[}8, 8{]} & 36.9K & 1.6K & 23.0$\times$ \\
layer1.1.conv1 & (64, 64, 3, 3) & {[}8, 8{]} & 36.9K & 1.6K & 23.0$\times$ \\
layer1.1.conv2 & (64, 64, 3, 3) & {[}8, 8{]} & 36.9K & 1.6K & 23.0$\times$ \\
layer2.0.conv1 & (128, 64, 3, 3) & {[}8, 8{]} & 73.7K & 2.11K & 34.9$\times$ \\
layer2.0.conv2 & (128, 128, 3, 3) & {[}16, 14{]} & 147K & 5.86K & 25.2$\times$ \\
layer2.1.conv1 & (128, 128, 3, 3) & {[}14, 13{]} & 147K & 5.09K & 28.9$\times$ \\
layer2.1.conv2 & (128, 128, 3, 3) & {[}11, 9{]} & 147K & 3.45K & 42.7$\times$ \\
layer3.0.conv1 & (256, 128, 3, 3) & {[}29, 29{]} & 295K & 18.7K & 15.8$\times$ \\
layer3.0.conv2 & (256, 256, 3, 3) & {[}62, 49{]} & 590K & 55.8K & 10.6$\times$ \\
layer3.1.conv1 & (256, 256, 3, 3) & {[}57, 55{]} & 590K & 56.9K & 10.4$\times$ \\
layer3.1.conv2 & (256, 256, 3, 3) & {[}39, 38{]} & 590K & 33K & 17.8$\times$ \\
layer4.0.conv1 & (512, 256, 3, 3) & {[}67, 66{]} & 1.18M & 91K & 13.0$\times$ \\
layer4.0.conv2 & (512, 512, 3, 3) & {[}98, 91{]} & 2.36M & 177K & 13.3$\times$ \\
layer4.1.conv1 & (512, 512, 3, 3) & {[}102, 94{]} & 2.36M & 187K & 12.6$\times$ \\
layer4.1.conv2 & (512, 512, 3, 3) & {[}20, 20{]} & 2.36M & 24.1K & 98.0$\times$ \\

\multicolumn{6}{c}{\textbf{32$\times$ Compression Ratio}} \\

layer1.0.conv1 & (64, 64, 3, 3) & {[}8, 8{]} & 36.9K & 1.6K & 23.0$\times$ \\
layer1.0.conv2 & (64, 64, 3, 3) & {[}8, 8{]} & 36.9K & 1.6K & 23.0$\times$ \\
layer1.1.conv1 & (64, 64, 3, 3) & {[}8, 8{]} & 36.9K & 1.6K & 23.0$\times$ \\
layer1.1.conv2 & (64, 64, 3, 3) & {[}8, 8{]} & 36.9K & 1.6K & 23.0$\times$ \\
layer2.0.conv1 & (128, 64, 3, 3) & {[}8, 8{]} & 73.7K & 2.11K & 34.9$\times$ \\
layer2.0.conv2 & (128, 128, 3, 3) & {[}8, 9{]} & 147K & 2.82K & 52.2$\times$ \\
layer2.1.conv1 & (128, 128, 3, 3) & {[}8, 8{]} & 147K & 2.62K & 56.2$\times$ \\
layer2.1.conv2 & (128, 128, 3, 3) & {[}8, 8{]} & 147K & 2.62K & 56.2$\times$ \\
layer3.0.conv1 & (256, 128, 3, 3) & {[}16, 17{]} & 295K & 8.72K & 33.8$\times$ \\
layer3.0.conv2 & (256, 256, 3, 3) & {[}35, 32{]} & 590K & 27.2K & 21.7$\times$ \\
layer3.1.conv1 & (256, 256, 3, 3) & {[}28, 30{]} & 590K & 22.4K & 26.3$\times$ \\
layer3.1.conv2 & (256, 256, 3, 3) & {[}16, 16{]} & 590K & 10.5K & 56.2$\times$ \\
layer4.0.conv1 & (512, 256, 3, 3) & {[}40, 43{]} & 1.18M & 47K & 25.1$\times$ \\
layer4.0.conv2 & (512, 512, 3, 3) & {[}55, 58{]} & 2.36M & 86.6K & 27.3$\times$ \\
layer4.1.conv1 & (512, 512, 3, 3) & {[}59, 57{]} & 2.36M & 89.7K & 26.3$\times$ \\
layer4.1.conv2 & (512, 512, 3, 3) & {[}12, 11{]} & 2.36M & 13K & 182.0$\times$ \\
\bottomrule
\end{tabular}
}
\label{table:resnet18_ranks}
\end{table*}

\newpage
\bibliography{aaai23.bib}